# FINDING A POSTERIOR DOMAIN PROBABILITY DISTRIBUTION BY SPECIFYING NONSPECIFIC EVIDENCE

JOHAN SCHUBERT
*Division of Information System Technology,
Department of Command and Control Warfare Technology,
National Defence Research Establishment,
S-172 90 Stockholm, SWEDEN*

This article is an extension of the results of two earlier articles. In [J. Schubert, "On nonspecific evidence", *Int. J. Intell. Syst.* **8** (1993) 711-725] we established within Dempster-Shafer theory a criterion function called the metaconflict function. With this criterion we can partition into subsets a set of several pieces of evidence with propositions that are weakly specified in the sense that it may be uncertain to which event a proposition is referring. In a second article [J. Schubert, "Specifying nonspecific evidence", in "Cluster-based specification techniques in Dempster-Shafer theory for an evidential intelligence analysis of multiple target tracks", Ph.D. Thesis, TRITA-NA-9410, Royal Institute of Technology, Stockholm, 1994, ISBN 91-7170-801-4] we not only found the most plausible subset for each piece of evidence, we also found the plausibility for every subset that this piece of evidence belongs to the subset. In this article we aim to find a posterior probability distribution regarding the number of subsets. We use the idea that each piece of evidence in a subset supports the existence of that subset to the degree that this piece of evidence supports anything at all. From this we can derive a bpa that is concerned with the question of how many subsets we have. That bpa can then be combined with a given prior domain probability distribution in order to obtain the sought-after posterior domain distribution.

*Keywords*: belief functions, Dempster-Shafer theory, evidential reasoning, evidence correlation, cluster analysis, posterior distribution.

## 1. Introduction

In two earlier articles [1, 2] we derived methods, within the framework of Dempster-Shafer theory [3-7], to handle pieces of evidence that are weakly specified in the sense that it may not be certain to which of several possible events a proposition is referring. When reasoning with such pieces of evidence we must avoid combining the pieces of evidence by mistake that refer to different events.

The methodology developed in these two articles was intended for a multiple-target tracking algorithm in an anti-submarine intelligence analysis system [1, 8-9]. In this application a sparse flow of intelligence reports arrives at the analysis system. These reports may originate from several different unconnected sensor systems. The reports carry a proposition about the occurrence of a submarine at a specified time and place, a probability of the truthfulness of the report and may contain additional information such as velocity, direction and type of submarine.

When there are several submarines we want to separate the intelligence reports into subsets according to which submarine they are referring to. We will then analyze the reports for each submarine separately. However, the intelligence reports are never labeled as to which submarine they are referring to. Thus, it is not possible to directly differentiate between two different submarines using two intelligence reports.

Instead we will use the conflict between the propositions of two intelligence reports as a probability that the two reports are referring to different submarines. This probability is the basis for separating intelligence reports into subsets.

The cause of the conflict can be non-firing sensors placed between the positions of the two reports, the required velocity to travel between the positions of the two reports at their respective times in relation to the assumed velocity of the submarines, etc.

The general idea is this. If we receive several pieces of evidence about several different and separate events and the pieces of evidence are mixed up, we want to sort all the pieces of evidence according to which event they are referring to. Thus, we partition the set of all pieces of evidence into subsets where each subset refers to a particular event. In Fig. 1 these subsets are denoted by $\chi_i$. Here, thirteen pieces of evidence are partitioned into four subsets. When the number of subsets is uncertain there will also be a "domain conflict" which is a conflict between the current number of subsets and domain knowledge. The partition is then simply an allocation of all pieces of evidence to the different events. Since these events do not have anything to do with each other, we will analyze them separately.

Now, if it is uncertain to which event some pieces of evidence is referring we have a problem. It could then be impossible to know directly if two different pieces of evidence are referring to the same event. We do not know if we should put them into the same subset or not. This problem is then a problem of organization. Evidence from different problems that we want to analyze are unfortunately mixed up and we are having some problem separating it.

To solve this problem, we can use the conflict in Dempster's rule when all pieces of evidence within a subset are combined, as an indication of whether these pieces of evidence belong together. The higher this conflict is, the less credible that they belong together.

Let us create an additional piece of evidence for each subset where the proposition of this additional piece of evidence states that this is not an "adequate partition". Let the proposition take a value equal to the conflict of the combination within the subset. These new pieces of evidence, one regarding each subset, reason about the partition of the original evidence. Just so we do not confuse them with the original evidence, let us call all this evidence "metalevel evidence" and let us say that its combination and the analysis of that combination take place on the "metalevel", Fig. 1.

In the combination of all pieces of metalevel evidence, one regarding each subset, we only receive support stating that this is not an "adequate partition". We may call this support a "metaconflict". The smaller this support is, the more credible the partition. Thus, the most credible partition is the one that minimizes the metaconflict.

In the first of these two articles we partitioned the set of all pieces of evidence into subsets, where each subset was representing a separate event. These subsets should then be handled separately by subsequent reasoning processes. This methodology was able to find the optimal partitioning of evidence among subsets as well as the optimal estimate of the number of subsets when our own domain knowledge regarding the actual number of subsets was uncertain.

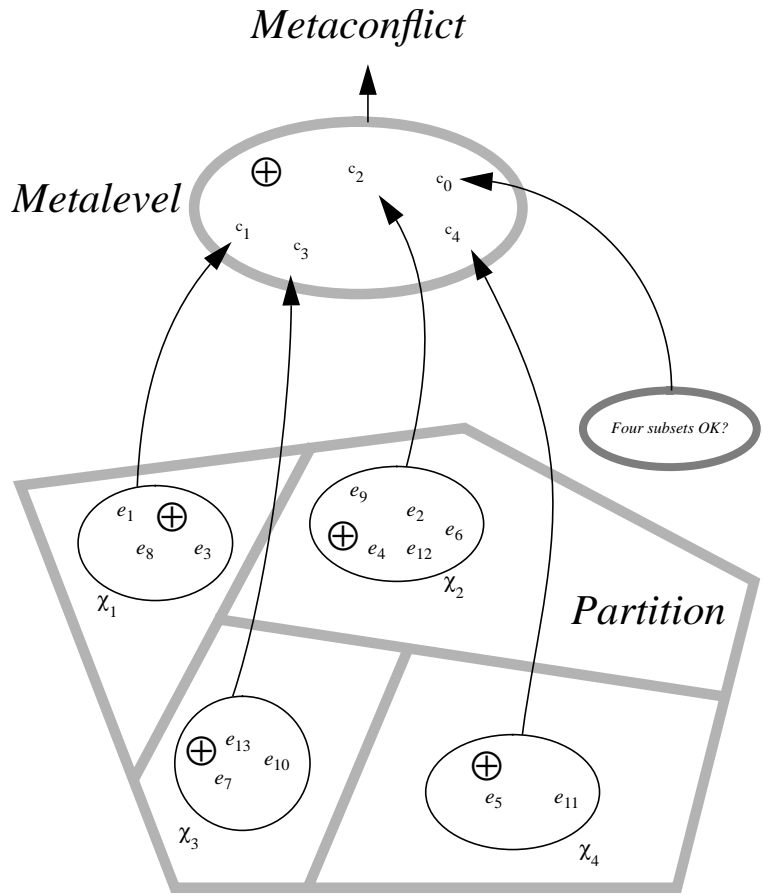

Fig. 1. The Conflict in each subset of the partition becomes a piece of evidence at the metalevel.

In the second article we found support regarding each piece of evidence and every subset, that the pieces of evidence does not belong to the subset. This support is used to specify each piece of evidence, in the sense that we find to which events the proposition might be referring, by calculating the belief and plausibility for each subset that this pieces of evidence belong to the subset. During this evidence specifying process we receive indications that some evidence might be false. Also, it became apparent that some pieces of evidence might not be so useful since they could belong to several different subsets. These pieces of evidence were discounted and were not allowed to strongly influence a subsequent reasoning process within a subset.

In this article we extend the work described in [1, 2] and aim to find a posterior probability distribution regarding the number of subsets by combining a given prior distribution with evidence regarding the number of subsets received from the evidence specifying process. We use the idea that each piece of evidence in a subset supports the existence of that subset to the degree that that piece of evidence supports anything at all. All pieces of evidence in each subset are combined and the resulting evidence is the total support for the subset. However, for every original piece of evidence in the subset we have a second piece of evidence, derived in [2], with a proposition that supports that this piece of evidence does not belong to the subset. If we have such support for every single piece of evidence in some subset, then this is also support that the subset is empty and that the proposition that the subset exist is false. Thus, in this case, we will discount the evidence that supports the existence of the subset. Such discounted pieces of evidence that support the existence of different subsets, one from each subset, are then combined.

From the resulting basic probability assignment (bpa) of that combination we can create a new bpa by exchanging each and every proposition. A proposition in the new bpa is then a statement about the existence of a minimal number of subsets where this number is the length of a conjunction of terms of the corresponding proposition in the previous bpa. Thus, where the previous bpa is concerned with the question of which subsets have support, the new bpa is concerned with the question of how many subsets are supported. The new bpa gives us some opinion, based only on the evidence specifying process, about the probability of different numbers of subsets.

In order to obtain the sought-after posterior domain probability distribution we combine the bpa from the evidence specifying process with the given prior distribution from the problem specification.

In Section 2 of this article we give a summary of the two previous articles [1, 2]. We investigate in Section 3 what domain relevant conclusions can be drawn from the evidence specifying process and then derive the posterior distribution. Finally, in Section 4, we give a detailed example.

## 2. Summary of articles [1, 2]

In this summary we will focus on results of the previous two articles that we need to derive a posterior domain probability distribution regarding the number of events. It will be derived by a combination of a given prior probability distribution and a bpa resulting from an evidence specification process [2] where we study the changes in conflict when we move a piece of evidence from one subset to another.

However, first we will learn how to separate several pieces of evidence based on their conflicts [1]. Since our pieces of evidence are weakly specified with respect to which events they are referring, it is impossible to directly separate the pieces of evidence based only on their proposition. The conflict in Dempster's rule measures the lack of compatibility between all pieces of evidence. Since pieces of evidence referring to different events tend to be more incompatible than pieces of evidence referring to the same event, it is an obvious choice as a distance measure in a cluster algorithm. The idea of using the conflict in Dempster's rule as distance measure between pieces of evidence was first suggested by Lowrance and Garvey [10].

### 2.1. *On nonspecific evidence [1]*

In [1] we established a criterion function of overall conflict called the metaconflict function. With this criterion we can partition evidence with weakly specified propositions into subsets, each subset representing a separate event. These events should be handled independently.

To make a separation of evidence possible, every proposition's action part must be supplemented with an event part describing to which event the proposition is referring. If the proposition is written as a conjunction of literals or disjunctions, then one literal or disjunction concerns which event the proposition is referring to. This is the event part. The remainder of the proposition is called the action part. An example from our earlier article illustrates the terminology:

> *Let us consider the burglaries of two bakers' shops at One and Two Baker Street, event 1 ($E_1$) and event 2 ($E_2$), i.e., the number of events is known to be two. One witness hands over a piece of evidence, specific with respect to event, with the proposition: "The burglar at One Baker Street," event part: $E_1$, "was probably brown haired (B)," action part: B. A second anonymous witness hands over a nonspecific piece of evidence with the proposition: "The burglar at Baker Street," event part: $E_1$, $E_2$, "might have been red haired (R)," action part: R. That is, for example:*

*evidence 1:*
 *proposition:*
  *action part: B*
  *event part: $E_1$:*
$m(B) = 0.8$
$m(\Theta) = 0.2$

*evidence 2:*
 *proposition:*
  *action part: R*
  *event part: $E_1, E_2$*
$m(R) = 0.4$
$m(\Theta) = 0.6$

### 2.1.1. *Separating nonspecific evidence*

We will have a conflict between two pieces of evidence in the same subset in two different situations. First, we have a conflict if the proposition action parts are conflicting regardless of the proposition event parts since they are presumed to be referring to the same event. Secondly, if the proposition event parts are conflicting then, regardless of the proposition action parts, we have a conflict with the presumption that they are referring to the same event.

The metaconflict used to partition the evidence is derived as the plausibility that the partitioning is correct when the conflict in each subset is viewed as a piece of metalevel evidence against the partitioning of the evidence, $\chi$, into the subsets, $\chi_i$. We have a simple frame of discernment on the metalevel $\Theta = \{\text{AdP}, \neg\text{AdP}\}$, where AdP is short for "adequate partition", and a bpa from each subset $\chi_i$ assigning support to a proposition against the partitioning:

$$m_{\chi_i}(\neg\text{AdP}) \triangleq \text{Conf}(\{e_j | e_j \in \chi_i\}),$$

$$m_{\chi_i}(\Theta) \triangleq 1 - \text{Conf}(\{e_j | e_j \in \chi_i\})$$

where $e_j$ is the *j*th piece of evidence and $\{e_j | e_j \in \chi_i\}$ is the evidence belonging to subset $\chi_i$ and $\text{Conf}(\cdot)$ is the conflict, *k*, in Dempster's rule. Also, we have a bpa concerning the domain resulting from a probability distribution about the number of subsets, *E*, conflicting with the actual current number of subsets, #$\chi$. This bpa also assigns support to a proposition against the partitioning:

$$m_D(\neg\text{AdP}) \triangleq \text{Conf}(\{E, \#\chi\}),$$

$$m_D(\Theta) \triangleq 1 - \text{Conf}(\{E, \#\chi\}).$$

The combination of these by Dempster's rule give us the following plausibility of the partitioning:

$$\text{Pls}(\text{AdP}) = (1 - m_D(\neg\text{AdP})) \cdot \prod_{i=1}^{r} (1 - m_{\chi_i}(\neg\text{AdP})).$$

The difference, one minus the plausibility of a partitioning, will be called the metaconflict of the partitioning.

### 2.1.2. *Metaconflict as a criterion function*

The metaconflict function can then be defined as:

DEFINITION. *Let the* metaconflict function,

$$Mcf(r, e_1, e_2, ..., e_n) \triangleq 1 - (1 - c_0) \cdot \prod_{i=1}^{r} (1 - c_i),$$

*be the conflict against a partitioning of n pieces of evidence of the set χ into r disjoint subsets $\chi_i$ where*

$$c_0 = \sum_{i \neq r} m(E_i)$$

*is the conflict between the hypothesis that there are r subsets and our prior belief about the number of subsets with $m(E_i)$ being the prior support given to the fact there are i subsets*

$$c_i = \sum_{\substack{I \\ \cap I = \emptyset}} \prod_{e_j^k \in I} m_j(e_j^k)$$

*is the conflict in subset i, where $\cap I$ is the intersection of all elements in I, $I = \{e_j^k | e_j \in \chi_i\}$ is a set of one focal element from the support function of each piece of evidence $e_j$ in $\chi_i$ and $e_j^k$ is the kth focal element of $e_j$.*

Thus, $|I| = |\chi_i|$ and

$$|\{I\}| = \prod_{e_j \in \chi_i} |e_j|$$

where $|e_j|$ is the number of focal elements of $e_j$.

We are here only considering the case where the function $m(\cdot)$ in the calculation of $c_0$ is a probability function.

Two theorems are derived to be used in the separation of evidence into subsets by an iterative minimization of the metaconflict function. By using these theorems we are able to reason about the optimal estimate of the number of events, when the actual number of events may be uncertain, as well as the optimal partition of nonspecific evidence for any fixed number of events. These two theorems will also be useful in a process for specifying pieces of evidence by observing changes in the metaconflict when moving a single piece of evidence between different subsets.

THEOREM 1. *For all j with $j < r$, if $m(E_j) < m(E_r)$ then min $Mcf(r,e_1,e_2,...,e_n) <$ min $Mcf(j,e_1,e_2,...,e_n)$.*

This theorem states that an optimal partitioning for *r* subsets is always better than the other solutions with fewer than *r* subsets if the basic probability assignment for *r* subsets is greater than the basic probability assignment for the fewer subsets.

THEOREM 2. *For all j, if min $Mcf(r, e_1, e_2, ..., e_n) < \sum_{i \neq j} m(E_i)$ then min $Mcf(r,e_1,e_2,...,e_n) <$ min $Mcf(j,e_1,e_2,...,e_n)$.*

Theorem 2 states that an optimal partitioning for some number of subsets is always better than other solutions for any other number of subsets when the domain part of the metaconflict function is greater than the total metaconflict of the present partitioning.

## 2.2. *Specifying nonspecific evidence [2]*

### 2.2.1. *Evidence about evidence*

A conflict in a subset $\chi_i$ is interpreted as a piece of metalevel evidence that there is at least one piece of evidence that does not belong to the subset;

$$m_{\chi_i}(\exists j . e_j \notin \chi_i) = c_i.$$

If a piece of evidence $e_q$ in $\chi_i$ is taken out from the subset the conflict $c_i$ in $\chi_i$ decreases to $c_i^*$. This decrease $c_i - c_i^*$ was interpreted as a piece of metalevel evidence indicating that $e_q$ does not belong to $\chi_i$, $m_{\Delta\chi_i}(e_q \notin \chi_i)$, and the remaining conflict $c_i^*$ is an other piece of metalevel evidence indicating that there is at least one other piece of evidence $e_j$, $j \neq q$, that does not belong to $\chi_i$ - $\{e_q\}$,

$$m_{\chi_i - \{e_q\}}(\exists j \neq q . e_j \notin (\chi_i - \{e_q\})) = c_i^*.$$

The unknown bpa, $m_{\Delta\chi_i}(e_q \notin \chi_i)$, was derived by stating that the belief that there is at least one piece of evidence that does not belong to $\chi_i$ should be equal, no matter whether that belief is based on the original piece of metalevel evidence $m_{\chi_i}(\exists j . e_j \notin \chi_i)$, before $e_q$ is taken out from $\chi_i$, or on a combination of the other two pieces of metalevel evidence $m_{\Delta\chi_i}(e_q \notin \chi_i)$ and $m_{\chi_i - \{e_q\}}(\exists j \neq q . e_j \notin (\chi_i - \{e_q\}))$, after $e_q$ is taken out from $\chi_i$, i.e.

$$\text{Bel}_{\chi_i}(\exists j . e_j \notin \chi_i) = \text{Bel}_{\Delta\chi_i \oplus (\chi_i - \{e_q\})}(\exists j . e_j \notin \chi_i).$$

where

$$\text{Bel}_{\chi_i}(\exists j . e_j \notin \chi_i) = c_i$$

and

$$\text{Bel}_{\Delta\chi_i \oplus (\chi_i - \{e_q\})}(\exists j. e_j \notin \chi_i) = c_i^* + m_{\Delta\chi_i}(e_q \notin \chi_i) \cdot [1 - c_i^*].$$

Thus, we derived a piece of metalevel evidence with a proposition stating that $e_q$ does not belong to $\chi_i$ from the variations in cluster conflict when $e_q$ was taken out from $\chi_i$:

$$m_{\Delta\chi_i}(e_q \notin \chi_i) = \frac{c_i - c_i^*}{1 - c_i^*}.$$

If $e_q$ after it is taken out from $\chi_i$ is brought into another subset $\chi_k$, its conflict will increase from $c_k$ to $c_k^*$. The increase in conflict when $e_q$ is brought into $\chi_k$ is interpreted as if there exists some evidence indicating that $e_q$ does not belong to $\chi_k + \{e_q\}$, i.e.

$$\forall k \neq i. m_{\Delta\chi_k}(e_q \notin (\chi_k + \{e_q\})) = \frac{c_k^* - c_k}{1 - c_k}.$$

When we take out a piece of evidence $e_q$ from subset $\chi_i$ and move it to some other subset we might have a changes in domain conflict. The domain conflict was interpreted as a piece of metalevel evidence that there exists at least one piece of evidence that does not belong to any of the $n$ first subsets, $n = |\chi|$, or if that particular piece of evidence was in a subset by itself, as a piece of metalevel evidence that it belongs to one of the other $n$-1 subsets. This indicate that the number of subsets is incorrect.

When $|\chi_i| > 1$ we may not only put a piece of evidence $e_q$ that we have taken out from $\chi_i$ into another already existing subset, we may also put $e_q$ into a new subset $\chi_{n+1}$ by itself. There is no change in the domain conflict when we take out $e_q$ from $\chi_i$ since $|\chi_i| > 1$, thus we may interpret the domain conflict as

$$m_\chi(\exists j \neq q \forall k \neq n+1. e_j \notin \chi_k) = c_0.$$

However, we will get an increase in domain conflict from $c_0$ to $c_0^*$ when we move $e_q$ to $\chi_{n+1}$. This increase is a piece of metalevel evidence indicating that $e_q$ does not belong to $\chi_{n+1}$, $m_{\Delta\chi}(e_q \notin \chi_{n+1})$, and the new domain conflict after $e_q$ is moved into $\chi_{n+1}$ is interpreted as

$$m_{\chi + \{\chi_{n+1}\}}(\exists j \forall k. e_j \notin \chi_k) = c_0^*.$$

We will derive $m_{\Delta\chi}(e_q \notin \chi_{n+1})$ by stating that

172  *J. Schubert*

$$\text{Bel}_{\chi + \{\chi_{n+1}\}}(\exists j \forall k. e_j \notin \chi_k) = \text{Bel}_{\Delta\chi \oplus \chi}(\exists j \forall k. e_j \notin \chi_k),$$

where

$$\text{Bel}_{\chi + \{\chi_{n+1}\}}(\exists j \forall k. e_j \notin \chi_k) = c_0^*$$

and

$$\text{Bel}_{\Delta\chi \oplus \chi}(\exists j \forall k. e_j \notin \chi_k) = c_0 + m_{\Delta\chi}(e_q \notin \chi_{n+1}) \cdot [1 - c_0].$$

Thus, we received

$$m_{\Delta\chi}(e_q \notin \chi_{n+1}) = \frac{c_0^* - c_0}{1 - c_0}$$

as the sought for piece of metalevel evidence, indicating that $e_q$ does not belong to $\chi_{n+1}$.

We will also receive a piece of evidence from domain conflict variations if $e_q$ is in a subset $\chi_i$ by itself and moved from $\chi_i$ to another already existing subset. In this case we may get either an increase or decrease in domain conflict. First, if the domain conflict decreases $c_0^* < c_0$ when we move $e_q$ out from $\chi_i$ this is interpreted as evidence that $e_q$ does not belongs to $\chi_i$,

$$m_{\Delta\chi}(e_q \notin \chi_i) = \frac{c_0 - c_0^*}{1 - c_0^*}.$$

Secondly, if we observe an increase in domain conflict $c_0^* > c_0$ we will interpret this as a new type of evidence, supporting the case that $e_q$ does belong to $\chi_i$;

$$m_{\Delta\chi}(e_q \in \chi_i) = \frac{c_0}{c_0^*}.$$

### 2.2.2. *Specifying evidence*

We may now combine the pieces of evidence from conflict variations and calculate the belief and plausibility for each subset that $e_q$ belongs to the subset. The belief for this will always be zero, except when $e_q \in \chi_i$, $|\chi_i| = 1$ and $c_0 < c_0^*$, since every proposition with this one exception states that $e_q$ does not belong to some subset.

When all pieces of evidence regarding $e_q$ are combined we have received support for a proposition stating that $e_q$ does not belong to any of the subsets and can not be put into a subset by itself. That proposition is false and its support is the conflict in Dempster's rule and an implication that this piece of evidence is false.

For the case when $e_q$ is in $\chi_i$ and $|\chi_i| > 1$ we combined all pieces of evidence regarding $e_q$ and receive a new basic probability assignment with

$$\forall \chi^* . m^*(e_q \notin (\vee \chi^*)) = \frac{1}{1-k} \cdot \prod_{\chi_j \in \chi^*} m(e_q \notin \chi_j) \cdot \prod_{\chi_j \in (\chi - \chi^*)} [1 - m(e_q \notin \chi_j)]$$

where $\chi^* \in 2^\chi$, $\chi = \{\chi_1, ..., \chi_{n+1}\}$ and $\vee \chi^*$ is the disjunction of all elements in $\chi^*$ and

$$k = \prod_{j=1}^{n+1} m(e_q \notin \chi_j).$$

This gave us plausibilities that $e_q$ belongs to a subset of

$$\forall k \neq n+1 . \mathrm{Pls}(e_q \in \chi_k) = \frac{1 - m(e_q \notin \chi_k)}{1 - \prod_{j=1}^{n+1} m(e_q \notin \chi_j)}$$

and

$$\mathrm{Pls}(e_q \in \chi_{n+1}) = \frac{1 - m(e_q \notin \chi_{n+1})}{1 - \prod_{j=1}^{n+1} m(e_q \notin \chi_j)}.$$

For the situation when $e_q \in \chi_i$, $|\chi_i| = 1$ and $c_0 > c_0^*$, the only change was that the domain conflict variation appeared in the $i^{\mathrm{th}}$ piece of evidence instead of the $n+1^{\mathrm{th}}$. This gave us a slight change in the calculation of plausibility. For subsets except $\chi_i$ we had

$$\forall k \neq i . \mathrm{Pls}(e_q \in \chi_k) = \frac{1 - m(e_q \notin \chi_k)}{1 - \prod_{j=1}^{n} m(e_q \notin \chi_j)}$$

and for $\chi_i$

$$\text{Pls}(e_q \in \chi_i) = \frac{1 - m(e_q \notin \chi_i)}{1 - \prod_{j=1}^{n} m(e_q \notin \chi_j)}.$$

When $e_q \in \chi_i$, $|\chi_i| = 1$ and $c_0 < c_0^*$ we did not receive any conflict in the combination of all pieces of evidence regarding $e_q$ since we had no evidence against the proposition that $e_q$ belonged to $\chi_i$. Furthermore, when we calculate belief and plausibility for any subset other than $\chi_i$ we have a zero belief in that $e_q$ belongs to $\chi_k$ but we receive a plausibility of

$$\forall k \neq i. \text{Pls}(e_q \in \chi_k) = [1 - m(e_q \in \chi_i)] \cdot [1 - m(e_q \notin \chi_k)]$$

and for $\chi_i$ we receive a belief of

$$\text{Bel}(e_q \in \chi_i) = m(e_q \in \chi_i) + [1 - m(e_q \in \chi_i)] \cdot \prod_{\chi_j \in \chi^{-i}} m(e_q \notin \chi_j)$$

where $\chi^{-i} = \chi - \{\chi_i\}$, $\chi = \{\chi_1, ..., \chi_n\}$ and a plausibility of one.

### 2.2.3. *Handling the falsity of evidence*

In combining all pieces of evidence regarding $e_q$ in Sec. 2.2.2 we received some support $k$ for the proposition that $e_q$ did not belong to any of the subsets. This is impossible and implies to a degree $k$ that $e_q$ is a false piece of evidence. If we had no indication as to the possible falsity of $e_q$ we would take no action, but if there existed such an indication we would pay ever less regard to this piece of evidence the higher the degree was that it is false and pay no attention to this piece of evidence when it is certainly false. This was done by discounting the evidence with its credibility $\alpha$,

$$m^{\%}(A_j) = \begin{cases} \alpha \cdot m(A_j), & A_j \neq \Theta, \\ 1 - \alpha + \alpha \cdot m(\Theta), & A_j = \Theta \end{cases}$$

where $A_j$ is $e_q$ or $\Theta$, $m(\cdot)$ the original piece of evidence, and where the credibility $\alpha$ is defined as one minus the support in the false proposition that $e_q$ does not belong to any subset and cannot be put in a subset by itself, i.e. one minus the conflict in Dempster's rule when combining all pieces of evidence regarding $e_q$;

$$\alpha \stackrel{\Delta}{=} 1 - m^*(e_q \notin (\ \vee \chi)) = 1 - k.$$

### 2.2.4. *Finding usable evidence*

If we plan to use $e_q$ in the reasoning process of some event, we must find the credibility that $e_q$ belongs to the subset in question and then discount this piece of evidence by its credibility.

Here we should note that each original piece of evidence can be used in the reasoning process of any subset that it belongs to with a plausibility above zero, given only that it is discounted to its credibility in belonging to the subset.

A piece of evidence that cannot possible belong to a subset $\chi_i$ has a credibility of zero and should be discounted entirely for that subset, while a piece of evidence which cannot possibly belong to any other subset $\chi_j$ and is without any support whatsoever against $\chi_i$ has a credibility of one and should not be discounted at all when used in the reasoning process for $\chi_i$. That is, the degree to which a piece of evidence can belong to a subset and no other subset corresponds to the importance it should be allowed to play in that subset.

We derived the credibility $\alpha_j$ of $e_q$ when $e_q$ is used in $\chi_j$ as

$$\alpha_j = [1 - \text{Bel}(e_q \in \chi_i)] \cdot \frac{[\text{Pls}(e_q \in \chi_j)]^2}{\sum_k \text{Pls}(e_q \in \chi_k)}, \qquad j \neq i,$$

$$\alpha_i = \text{Bel}(e_q \in \chi_i) + [1 - \text{Bel}(e_q \in \chi_i)] \cdot \frac{[\text{Pls}(e_q \in \chi_i)]^2}{\sum_k \text{Pls}(e_q \in \chi_k)},$$

where $\text{Bel}(e_q \in \chi_i)$ is equal to zero except when $e_q \in \chi_i$, $|\chi_i| = 1$ and $c_0 < c_0^*$. This gave us a final discounted bpa as

$$m^{\%\%}{}_i(A_j) = \begin{cases} \alpha_i \cdot m^{\%}(A_j), & A_j \neq \Theta \\ 1 - \alpha_i + \alpha_i \cdot m^{\%}(\Theta), & A_j = \Theta \end{cases}.$$

## 3. Deriving a posterior domain probability distribution

### 3.1. *Evidence about subsets*

We use the idea that each piece of evidence in a subset supports the existence of that subset to the degree that that piece of evidence supports anything at all. For a subset $\chi_i$, each single piece of evidence we have is discounted for its degree of falsity and its degree of credibility in belonging to $\chi_i$, $m_q^{\%\%}{}_i$. All discounted pieces of evidence in $\chi_i$ are then combined. The value of all $m_q^{\%\%}{}_i$'s were derived in [2] from the $m_q$'s by the specifying process. The degree to which the bpa resulting from this combination supports anything at all other than the entire frame is the degree to which these pieces of evidence taken together supports the existence of $\chi_i$, i.e. that $\chi_i$ is a nonempty subset that belongs to $\chi$. Thus, we have

$$m_{\chi_i}(\chi_i \in \chi) = 1 - \frac{1}{1-k} \cdot \prod_q m_q^{\%\%i}(\Theta),$$

$$m_{\chi_i}(\Theta) = \frac{1}{1-k} \cdot \prod_q m_q^{\%\%i}(\Theta)$$

where $k$ is the conflict in Dempster's rule when combining all $m_q^{\%\%i}$.

For every piece of evidence we have some support in favor of it not belonging to the subset. To the degree that this is fulfilled for all pieces of evidence in $\chi_i$ it supports the case that none of the pieces of evidence that could belong to $\chi_i$ actually did so. That is, it is support for the case that the proposition $\chi_i \in \chi$ is false. Thus, we would like to discount the just derived pieces of evidence as

$$m_{\chi_i}^{\%}(\chi_i \in \chi) = \alpha_i \cdot m_{\chi_i}(\chi_i \in \chi),$$

$$m_{\chi_i}^{\%}(\Theta) = 1 - \alpha_i + \alpha_i \cdot m_{\chi_i}(\Theta)$$

where

$$\alpha_i = \begin{cases} 1, & |\chi_i| = 1, c_0 < c_0^* \\ 1 - f_i \cdot g_i \cdot h_i, & \text{otherwise} \end{cases}$$

with

$$f_i = \prod_{q|\, e_q \in \chi_i} m(e_q \notin \chi_i),\ [\,(|\chi_i| > 1) \vee (|\chi_i| = 1, c_0 > c_0^*)\,],$$

$$g_i = \begin{cases} \prod_{q|\, e_q \in \chi_j,\, j \neq i} m(e_q \notin \chi_i), & (|\chi_i| > 0,\ [\,(|\chi_j| > 1) \vee (|\chi_j| = 1, c_0 > c_0^*)\,]) \\ \prod_{q|\, e_q \in \chi_j,\, j \neq n+1} m(e_q \notin \chi_{n+1}), & |\chi_{n+1}| = 0, |\chi_j| > 1 \end{cases}$$

and

$$h_i = \prod_{q|\, e_q \in \chi_j,\, j \neq i} [\,1 - (1 - m(e_q \notin \chi_i)) \cdot (1 - m(e_q \in \chi_j))\,],\ |\chi_i| > 0, |\chi_j| = 1, c_0 < c_0^*$$

where $f_i \cdot g_i \cdot h_i$ is the support that $\chi_i$ is empty, i.e. support that $\chi_i$ does not exist. Here we have, from [2], (for all $i$ when $e_q \in \chi_i$ we have $m(e_q \notin \chi_j) = \ldots$),

$$\forall i, e_q \in \chi_i . m(e_q \notin \chi_j) = \begin{cases} \dfrac{c_0^* - c_0}{1 - c_0}, j = n + 1, |\chi_i| > 1 \\[6pt] \dfrac{c_i - c_i^*}{1 - c_i^*}, j = i, |\chi_i| > 1 \\[6pt] \dfrac{c_0 - c_0^*}{1 - c_0^*}, j = i, |\chi_i| = 1, c_0 > c_0^* \\[6pt] \dfrac{c_j^* - c_j}{1 - c_j}, \text{otherwise} \end{cases}$$

and

$$\forall i, e_q \in \chi_i . m(e_q \in \chi_i) = \dfrac{c_0}{c_0^*}, |\chi_i| = 1, c_0 < c_0^*$$

where $c_i$ and $c_i^*$ are conflicts in subset $\chi_i$ before and after $e_q$ is taken out from the subset, $c_j^*$ and $c_j$ are conflict in a subset $\chi_j$, $j \neq i$, before and after $e_q$ was brought into the subset, and $c_0$ and $c_0^*$ are domain conflicts before and after $e_q$ was brought either from a subset with several pieces of evidence into a new subset or, if it is in a subset by itself, from this subset into one of the other already existing subsets.

### 3.2. *Evidence about the number of subsets*

The discounted pieces of evidence $m_{\chi_i}^\%$, one from each subset, are then combined. The resulting bpa will then have focal elements that supports propositions such as

$$(\chi_1 \in \chi) \wedge (\chi_3 \in \chi) \wedge (\chi_4 \in \chi).$$

We have

$$m_\chi^\%((\wedge \chi^*) \in \chi) = \prod_{i | (\chi_i \in \chi^*)} m_{\chi_i}^\%(\chi_i \in \chi) \cdot \prod_{j | (\chi_j \notin \chi^*)} m_{\chi_j}^\%(\Theta),$$

$$m_\chi^\%(\Theta) = \prod_{i=1}^n m_{\chi_i}^\%(\Theta).$$

From this we can create a new bpa by exchanging all propositions in the previous bpa that are conjunctions of $r$ terms for one proposition in the new bpa that is on the form $|\chi| \geq r$. The sum of probability of all conjunctions of length $r$ in the previous bpa is then awarded the focal element in the new bpa which supports the proposition that $|\chi| \geq r$;

$$m_\chi(|\chi| \geq r) = \sum_{\chi^* \mid |\chi^*| = r} m_\chi^\%((\ \wedge \chi^*) \in \chi),$$

$$m_\chi(\Theta) = m_\chi^\%(\Theta)$$

where $\chi^* \in 2^\chi$ and $\chi = \{\chi_1, \chi_2, ..., \chi_n\}$.

A proposition in the new bpa is then a statement about the existence of some minimal number of subsets and its bpa taken as a whole gives us an opinion about the probability of different numbers of subsets.

### 3.3. *Combining the evidence with a prior distribution*

This newly created bpa can now be combined with our prior probability distribution, $m(\cdot)$, from the problem specification, to yield the demanded posterior probability distribution, $m^*(\cdot)$. We get

$$m^*(E_i) = \frac{1}{1-k} \cdot m(E_i) \cdot \left( m_\chi(\Theta) + \sum_{j=1}^{i} m_\chi(|\chi| \geq j) \right)$$

where

$$k = \sum_{i=0}^{n-1} \sum_{j=i+1}^{n} m(E_i) \cdot m_\chi(|\chi| \geq j)$$

is the conflict in that final combination.

Thus, by viewing each piece of evidence in a subset as support for the existence of that subset we were able to derive a bpa, concerned with the question of how many subsets we have, which we could combine with our prior domain probability distribution in order to obtain the sought-after posterior domain probability distribution.

### 4. An Example

In our first article [1] we described a problem involving two possible burglaries. In this example we had evidence weakly specified in the sense that it is uncertain to which possible burglary the propositions are referring. The problem we were facing was described as follows:

> *Assume that a baker's shop at One Baker Street has been burglarized, event* 1. *Let there also be some indication that a baker's shop across the street, at Two Baker Street, might have been burglarized, although no burglary has been reported, event* 2. *An* *experienced investigator estimates that a burglary has taken place at Two Baker Street with a probability of* 0.4. *We have received the following pieces of evidence. A credible witness reports that "a brown-haired man who is not an employee at the baker's shop committed the burglary at One Baker Street," evidence* 1. *An anonymous witness, not being aware that there might be two burglaries, has reported "a brown-haired man who works at the baker's shop committed the burglary at Baker Street," evidence* 2. *Thirdly, a witness reports having seen "a suspicious-looking red-haired man in the baker's shop at Two Baker Street," evidence* 3. *Finally, we have a fourth witness, this witness, also anonymous and not being aware of the possibility of two burglaries, reporting that the burglar at the Baker Street baker's shop was a brown-haired man. That is, for example:*

*evidence* 1:
  *proposition:*
    *action part: BO*
    *event part:* $E_1$:
$m(BO) = 0.8$
$m(\Theta) = 0.2$

*evidence* 2:
  *proposition:*
    *action part: BI*
    *event part:* $E_1, E_2$
$m(BI) = 0.7$
$m(\Theta) = 0.3$

*evidence* 3:
  *proposition:*
    *action part: R*
    *event part:* $E_2$:
$m(R) = 0.6$
$m(\Theta) = 0.4$

*evidence* 4:
  *proposition:*
    *action part: B*
    *event part:* $E_1, E_2$
$m(B) = 0.5$
$m(\Theta) = 0.5$

*domain probability distribution:*

$$m(E_i) = \begin{cases} 0.6, & i = 1 \\ 0.4, & i = 2 \\ 0, & i \neq 1, 2 \end{cases}.$$

All pieces of evidence were originally put into one subset, $\chi_1$. By minimizing the metaconflict function it was found best to partition the evidence into two subsets. The minimum of the metaconflict function was found when evidence one and four were moved from $\chi_1$ into $\chi_2$ while evidence two and three remained in $\chi_1$. This gave us a conflict in $\chi_1$ of $c_1 = 0.42$, in $\chi_2$ of $c_2 = 0$, and a domain conflict of $c_0 = 0.6$.

In our second article [2] we studied variations in the cluster conflict when a piece of evidence is moved from one subset to another, or put into a new subset by itself. Starting with $e_1$ we found that if $e_1$ in $\chi_2$ is moved out from $\chi_2$ the conflict remains at zero, $c_2^* = 0$. If $e_1$ then is moved into $\chi_1$ its conflict increased to $c_1^* = 0.788$, but if $e_1$ is instead put into a subset by itself, $\chi_3$, we will have a domain conflict of one, $c_0^* = 1$. By the formulas of [2] we received three bpa's regarding $e_1$:

$$m(e_1 \notin \chi_1) = \frac{c_1^* - c_1}{1 - c_1} = 0.634, \qquad m(e_1 \notin \chi_2) = \frac{c_2 - c_2^*}{1 - c_2^*} = 0$$

and

$$m(e_1 \notin \chi_3) = \frac{c_0^* - c_0}{1 - c_0} = 1,$$

with the remainder in each case awarded to the entire frame. We received for the other three pieces of evidence by the same formulas:

$$m(e_2 \notin \chi_i) = \begin{cases} 0.42, i = 1 \\ 0.56, i = 2 \\ 1, i = 3 \end{cases}, \qquad m(e_3 \notin \chi_i) = \begin{cases} 0.42, i = 1 \\ 0.54, i = 2 \\ 1, i = 3 \end{cases},$$

$$m(e_4 \notin \chi_i) = \begin{cases} 0.155, i = 1 \\ 0, i = 2 \\ 1, i = 3 \end{cases}.$$

In each case the remainder was awarded to the entire frame.

When the three bpa's regarding where a particular piece of evidence might belong were combined, a conflict was received for $e_2$ and $e_3$, but not for $e_1$ and $e_4$. Thus, there is no indication from this combination that $e_1$ and $e_4$ might be false. For the second and third piece of evidence a conflict of 0.2352 and 0.2268 was received, respectively. This is their degrees of falsity. Evidence $e_2$ and $e_3$ were then discounted to their respective degrees of credibility $\alpha = 1 - k$, i.e. 0.7648 and 0.7732:

$$m^{\%}(A_j) = \begin{cases} \alpha \cdot m(A_j), & A_j \neq \Theta \\ 1 - \alpha + \alpha \cdot m(\Theta), & A_j = \Theta \end{cases}.$$

This gave us

$m_1^{\%}(BO) = 0.8 \qquad m_2^{\%}(BI) = 0.5354 \qquad m_3^{\%}(R) = 0.4639 \qquad m_4^{\%}(B) = 0.5$

$m_1^{\%}(\Theta) = 0.2 \qquad m_2^{\%}(\Theta) = 0.4646 \qquad m_3^{\%}(\Theta) = 0.5361 \qquad m_4^{\%}(\Theta) = 0.5.$

Since all four pieces of evidence can belong to either of the two subsets it will always be uncertain if it belongs to a particular subset in question. In order to justify the use of a piece of evidence in some subset we must find the credibility that it belongs to the subset and discount it to its credibility. That is, an individual discounting is made for each subset and piece of evidence according to how credible it is that the piece of evidence belongs to the subset.

The credibility that $e_1$ belongs to $\chi_1$ is

$$\alpha_1 = \frac{(\text{Pls}(e_1 \in \chi_1))^2}{\sum_{j=1}^{2} \text{Pls}(e_1 \in \chi_j)} = \frac{0.366^2}{0.366 + 1} = 0.0981$$

where

$$\text{Pls}(e_1 \in \chi_1) = \frac{1 - m(e_1 \notin \chi_1)}{1 - m(e_1 \notin \chi_1) \cdot m(e_1 \notin \chi_2) \cdot m(e_1 \notin \chi_3)} = 1 - 0.634 = 0.366,$$

$$\text{Pls}(e_1 \in \chi_2) = \frac{1 - m(e_1 \notin \chi_2)}{1 - m(e_1 \notin \chi_1) \cdot m(e_1 \notin \chi_2) \cdot m(e_1 \notin \chi_3)} = 1.$$

and that $e_1$ belongs to $\chi_2$

$$\alpha_2 = \frac{(\text{Pls}(e_1 \in \chi_2))^2}{\sum_{j=1}^{2} \text{Pls}(e_1 \in \chi_j)} = \frac{1}{0.366 + 1} = 0.7321.$$

For the other three pieces of evidence we get: $e_2$: $\alpha_1 = 0.4310$, $\alpha_2 = 0.2480$, $e_3$: $\alpha_1 = 0.4182$, $\alpha_2 = 0.2632$, and for $e_4$: $\alpha_1 = 0.3870$, $\alpha_2 = 0.5420$.

Discounting the four pieces of evidence to their credibility of belonging to $\chi_1$ and $\chi_2$, respectively, we found:

$$m_1^{\%\%\,1}(BO) = 0.0784 \qquad m_1^{\%\%\,2}(BO) = 0.5856$$
$$m_1^{\%\%\,1}(\Theta) = 0.9216 \qquad m_1^{\%\%\,2}(\Theta) = 0.4144,$$

$$m_2^{\%\%\,1}(BI) = 0.2308 \qquad m_2^{\%\%\,2}(BI) = 0.1328$$
$$m_2^{\%\%\,1}(\Theta) = 0.7692 \qquad m_2^{\%\%\,2}(\Theta) = 0.8672,$$

$$m_3^{\%\%\,1}(R) = 0.1940 \qquad m_3^{\%\%\,2}(R) = 0.1221$$
$$m_3^{\%\%\,1}(\Theta) = 0.8060 \qquad m_3^{\%\%\,2}(\Theta) = 0.8779$$

and

$$m_4^{\%\%1}(B) = 0.1935 \qquad m_4^{\%\%2}(B) = 0.2710$$
$$m_4^{\%\%1}(\Theta) = 0.8065 \qquad m_4^{\%\%2}(\Theta) = 0.7290.$$

These results were derived in [2].

Starting with these results we begin the work to find a posterior probability distribution for the number of subsets.

By using the idea that each piece of evidence in a subset supports the existence of that subset to the degree that the piece of evidence supports anything at all, we calculate the support in our two subsets as

$$m_{\chi_1}(\chi_1 \in \chi) = 1 - \prod_q m_q^{\%\%1}(\Theta) = 1 - m_1^{\%\%1}(\Theta) \cdot m_2^{\%\%1}(\Theta) \cdot m_3^{\%\%1}(\Theta) \cdot m_4^{\%\%1}(\Theta) = 0.4893,$$
$$m_{\chi_1}(\Theta) = \prod_q m_q^{\%\%1}(\Theta) = m_1^{\%\%1}(\Theta) \cdot m_2^{\%\%1}(\Theta) \cdot m_3^{\%\%1}(\Theta) \cdot m_4^{\%\%1}(\Theta) = 0.5107$$

and

$$m_{\chi_2}(\chi_2 \in \chi) = 1 - \prod_q m_q^{\%\%2}(\Theta) = 1 - m_1^{\%\%2}(\Theta) \cdot m_2^{\%\%2}(\Theta) \cdot m_3^{\%\%2}(\Theta) \cdot m_4^{\%\%2}(\Theta) = 0.7268,$$
$$m_{\chi_2}(\Theta) = \prod_q m_q^{\%\%2}(\Theta) = m_1^{\%\%2}(\Theta) \cdot m_2^{\%\%2}(\Theta) \cdot m_3^{\%\%2}(\Theta) \cdot m_4^{\%\%2}(\Theta) = 0.2732.$$

If we have support for every single piece of evidence in some subset in favor of that the piece of evidence does not belong to the subset, then this is also support that the proposition $\chi_i \in \chi$ is false. In this case none of the pieces of evidence that could belong to the subset actually did so and the subset was derived by mistake. Thus, we will discount the just derived pieces of evidence that support the existence of the subsets for this possibility.

There is some evidence against the first subset, yielding a credibility for that subset of less than one

$$\alpha_1 = 1 - \prod_{q|\, e_q \in \chi_1} m(e_q \in \chi_1) \cdot \prod_{q|\, e_q \in \chi_j,\, j \notin 1} m(e_q \in \chi_1) = 0.9826,$$
$$\alpha_2 = 1 - \prod_{q|\, e_q \in \chi_2} m(e_q \in \chi_2) \cdot \prod_{q|\, e_q \in \chi_j,\, j \notin 2} m(e_q \in \chi_2) = 1.$$

We then discount the two bpa's that support the existence of the subsets to their respective credibility and receive

$$m_{\chi_1}^\%(\chi_1 \in \chi) = \alpha \cdot m_{\chi_1}(\chi_1 \in \chi) = 0.4808,$$

$$m_{\chi_1}^\%(\Theta) = 1 - \alpha - \alpha \cdot m_{\chi_1}(\Theta) = 0.5192$$

for the first subset and

$$m_{\chi_2}^\%(\chi_2 \in \chi) = \alpha \cdot m_{\chi_1}(\chi_2 \in \chi) = 0.7268,$$

$$m_{\chi_2}^\%(\Theta) = 1 - \alpha - \alpha \cdot m_{\chi_1}(\Theta) = 0.2732$$

for the second subset. If we then combine these two bpa's we receive

$$m_\chi^\%((\ \wedge \{\chi_1, \chi_2\}) \in \chi) = \prod_{i|\ (\chi_i \in \{\chi_1, \chi_2\})} m_{\chi_i}^\%(\chi_i \in \chi) \cdot \prod_{j|\ (\chi_j \notin \{\chi_1, \chi_2\})} m_{\chi_j}^\%(\Theta)$$

$$= m_{\chi_1}^\%(\chi_1 \in \chi) \cdot m_{\chi_2}^\%(\chi_2 \in \chi) = 0.3494,$$

$$m_\chi^\%((\ \wedge \{\chi_1\}) \in \chi) = \prod_{i|\ (\chi_i \in \{\chi_1\})} m_{\chi_i}^\%(\chi_i \in \chi) \cdot \prod_{j|\ (\chi_j \notin \{\chi_1\})} m_{\chi_j}^\%(\Theta)$$

$$= m_{\chi_1}^\%(\chi_1 \in \chi) \cdot m_{\chi_2}^\%(\Theta) = 0.1314,$$

$$m_\chi^\%((\ \wedge \{\chi_2\}) \in \chi) = \prod_{i|\ (\chi_i \in \{\chi_2\})} m_{\chi_i}^\%(\chi_i \in \chi) \cdot \prod_{j|\ (\chi_j \notin \{\chi_2\})} m_{\chi_j}^\%(\Theta)$$

$$= m_{\chi_2}^\%(\chi_2 \in \chi) \cdot m_{\chi_1}^\%(\Theta) = 0.3774,$$

$$m_\chi^\%(\Theta) = \prod_{i=1}^{2} m_{\chi_i}^\%(\Theta) = m_{\chi_1}^\%(\Theta) \cdot m_{\chi_2}^\%(\Theta) = 0.1418.$$

Given this result we create a new and final bpa by exchanging the focal elements of this bpa. Where the previous bpa is concerned with the question of which subsets have support, the new bpa is concerned with the question of how many subsets are supported. Thus, the new bpa gives us an opinion, based only on the result of the evidence specifying process, about the probability of different number of subsets. We have

$$m_\chi(|\chi| \geq 2) = \sum_{\chi^*|\ |\chi^*| = 2} m_\chi^\%((\ \wedge \chi^*) \in \chi) = m_\chi^\%((\ \wedge \{\chi_1, \chi_2\}) \in \chi) = 0.3494,$$

$$m_\chi(|\chi| \geq 1) = \sum_{\chi^*|\ |\chi^*| = 1} m_\chi^\%((\ \wedge \chi^*) \in \chi) = m_\chi^\%((\ \wedge \{\chi_1\}) \in \chi) + m_\chi^\%((\ \wedge \{\chi_2\}) \in \chi)$$

$$= 0.5087,$$

$$m_\chi(\Theta) = m_\chi^\%(\Theta) = 0.1418.$$

To conclude the analysis we combine this final bpa, from the evidence specifying process, with the given prior domain probability distribution from the problem specification,

$$m(E_i) = \begin{cases} 0.6, & i = 1 \\ 0.4, & i = 2 \\ 0, & \text{otherwise} \end{cases},$$

in order to receive the sought-after posterior domain distribution as the bpa of that combination. When doing this we receive a conflict of

$$k = \sum_{i=0}^{1} \sum_{j=i+1}^{2} m(E_i) \cdot m_\chi(|\chi| \geq j) = m(E_0) \cdot m_\chi(|\chi| \geq 1) + m(E_0) \cdot m_\chi(|\chi| \geq 1)$$
$$+ m(E_1) \cdot m_\chi(|\chi| \geq 2) = 0.2097$$

and obtain

$$m^*(E_2) = \frac{1}{1-k} \cdot m(E_2) \cdot \left(m_\chi(\Theta) + \sum_{j=1}^{2} m_\chi(|\chi| \geq j)\right)$$
$$= \frac{1}{1-k} \cdot m(E_2) \cdot (m_\chi(\Theta) + m_\chi(|\chi| \geq 1) + m_\chi(|\chi| \geq 2)) = 0.5061,$$

$$m^*(E_1) = \frac{1}{1-k} \cdot m(E_1) \cdot \left(m_\chi(\Theta) + \sum_{j=1}^{1} m_\chi(|\chi| \geq j)\right)$$
$$= \frac{1}{1-k} \cdot m(E_2) \cdot (m_\chi(\Theta) + m_\chi(|\chi| \geq 1)) = 0.4939,$$

$$m^*(E_i) = 0, \text{ otherwise}$$

as the posterior domain probability distribution. We find from the posterior distribution that the alternative with two events is slightly preferable to the one-event alternative.

## 5.  Conclusions

We have shown that it is possible to derive a posterior domain probability distribution from the reasoning process of specifying nonspecific evidence. This was done by viewing each piece of evidence in a subset as support for the existence of that subset. Based on this, we were able to find support for different number of subsets. Combined with a given prior distribution that yielded the sought-after posterior distribution.

The methodology described in this article builds on the work to partition the set of all pieces of evidence by minimizing a criterion function of overall conflict that was established within Dempster-Shafer theory [1] and also on the work of specifying evidence by studying changes in the conflict when a piece of evidence was moved from one subset to another [2].

## Acknowledgments

I would like to thank Stefan Arnborg, Ulla Bergsten, and Per Svensson for their helpful comments regarding this article.

## References

1. J. Schubert, "On nonspecific evidence", *Int. J. Intell. Syst.* **8** (1993) 711-725.
2. J. Schubert, "Specifying nonspecific evidence", in "Cluster-based specification techniques in Dempster-Shafer theory for an evidential intelligence analysis of multiple target tracks", Ph.D. Thesis, TRITA-NA-9410, Royal Institute of Technology, Stockholm, 1994, ISBN 91-7170-801-4.
3. A.P. Dempster, "A generalization of Bayesian inference", *J. R. Stat. Soc. Ser. B* **30** (1968) 205-247.
4. G. Shafer, *A Mathematical Theory of Evidence* (Princeton University, Princeton, 1976).
5. G. Shafer, "Perspectives on the theory and practice of belief functions", *Int. J. Approx. Reasoning* **4** (1990) 323-362.
6. G. Shafer, "Rejoinders to comments on 'Perspectives on the theory and practice of belief functions'", *Int. J. Approx. Reasoning* **6** (1992) 445-480.
7. R. Y. Yager, M. Fedrizzi, and J. Kacprzyk, *Advances in the Dempster-Shafer Theory of Evidence* (John Wiley & Sons, New York, 1994).
8. J. Schubert, "Cluster-based specification techniques in Dempster-Shafer theory", in *Symbolic and Quantitative Approaches to Reasoning and Uncertainty, Proc. 3rd European Conf. on Symbolic and Quantitative Approaches to Reasoning and Uncertainty,* eds. C. Froidevaux and J. Kohlas (Springer, Berlin, 1995), to appear.
9. U. Bergsten and J. Schubert, "Dempster's rule for evidence ordered in a complete directed acyclic graph", *Int. J. Approx. Reasoning* **9** (1993) 37-73.
10. J. D. Lowrance and T. D. Garvey, "Evidential reasoning: An implementation for multisensor integration", Technical Note 307, SRI International, Menlo Park, CA, 1983.